\documentclass[lettersize,journal]{IEEEtran}
\usepackage{amsmath,amsfonts}
\pdfoutput=1
\usepackage{algorithmic}
\usepackage{array}
\usepackage[caption=false,font=normalsize,labelfont=sf,textfont=sf]{subfig}
\usepackage{textcomp}
\usepackage{stfloats}
\usepackage{url}
\usepackage{verbatim}
\usepackage{graphicx}
\usepackage{amssymb}
\usepackage{booktabs}
\usepackage{multirow}
\usepackage{colortbl}
\usepackage[colorlinks,linkcolor=blue]{hyperref}
\usepackage{graphicx}
\usepackage{float} 
\definecolor{mygray}{gray}{.9}
\definecolor{mypink}{rgb}{.75,1,.75}
\definecolor{mycyan}{cmyk}{.3,0,0,0}
\def\BibTeX{{\rm B\kern-.05em{\sc i\kern-.025em b}\kern-.08em
    T\kern-.1667em\lower.7ex\hbox{E}\kern-.125emX}}
\usepackage{balance}
\begin{document}
\title{CAMO-MOT: Combined Appearance-Motion Optimization for 3D Multi-Object Tracking with Camera-LiDAR Fusion}

\author{\IEEEauthorblockN{
Li Wang, 
Xinyu Zhang,
Wenyuan Qin,
Xiaoyu Li,
Lei Yang,
Zhiwei Li,
Lei Zhu,
Hong Wang,
Jun Li, and
Huaping Liu
}

\thanks{This work was supported by the National High Technology Research and Development Program of China under Grant No. 2018YFE0204300, the National Natural Science Foundation of China under Grant No. U1964203, and the Guoqiang Research Institute Project (2019GQG1010), Meituan and Tsinghua University-Didi Joint Research Center for Future Mobility, the China Postdoctoral Science Foundation (No. 2021M691780), and State Key Laboratory of Robotics and Systems (HIT) (SKLRS-2022-KF-12). (\emph{Corresponding author: Xinyu Zhang.})

Li Wang, Xinyu Zhang, Wenyuan Qin, Xiaoyu Li, Lei Yang, Zhiwei Li, Hong Wang, and Jun Li are with the State Key Laboratory of Automotive Safety and Energy, and the School of Vehicle and Mobility, Tsinghua University, Beijing 100084, China (e-mail: wangli\_thu@mail.tsinghua.edu.cn; xyzhang@tsinghua.edu.cn; qinwenyuan1996@163.com; lixiaoyu200010@ 163.com; yanglei20@mails.tsinghua.edu.cn; lizhiwei713818@163.com; wangh@tsinghua.edu.cn; lj19580324@126.com).

Lei Zhu is with Mogo Auto Intelligence and Telemetics Information Technology Co. Ltd.

Huaping Liu is with the State Key Laboratory of Intelligent Technology and the Systems and Department of Computer Science and Technology, Tsinghua University, Beijing 100084, China (e-mail: hpliu@tsinghua.edu.cn).
}}


\maketitle

\begin{abstract}
3D Multi-object tracking (MOT) ensures consistency during continuous dynamic detection, conducive to subsequent motion planning and navigation tasks in autonomous driving. However, camera-based methods suffer in the case of occlusions and it can be challenging to accurately track the irregular motion of objects for LiDAR-based methods. Some fusion methods work well but do not consider the untrustworthy issue of appearance features under occlusion. At the same time, the false detection problem also significantly affects tracking. As such, we propose a novel camera-LiDAR fusion 3D MOT framework based on the Combined Appearance-Motion Optimization (CAMO-MOT), which uses both camera and LiDAR data and significantly reduces tracking failures caused by occlusion and false detection. For occlusion problems, we are the first to propose an occlusion head to select the best object appearance features multiple times effectively, reducing the influence of occlusions. To decrease the impact of false detection in tracking, we design a motion cost matrix based on confidence scores which improve the positioning and object prediction accuracy in 3D space. As existing multi-object tracking methods only consider a single category, we also propose to build a multi-category loss to implement multi-object tracking in multi-category scenes. A series of validation experiments are conducted on the KITTI and nuScenes tracking benchmarks. Our proposed method achieves state-of-the-art performance and the lowest identity switches (IDS) value (23 for Car and 137 for Pedestrian) among all multi-modal MOT methods on the KITTI test dataset. And our proposed method achieves state-of-the-art performance among all algorithms on the nuScenes test dataset with 75.3\% AMOTA. 
\end{abstract}


\section{Introduction}
\begin{figure}[t]
\begin{center}
   \includegraphics[width=1\linewidth,height=0.4\textwidth]{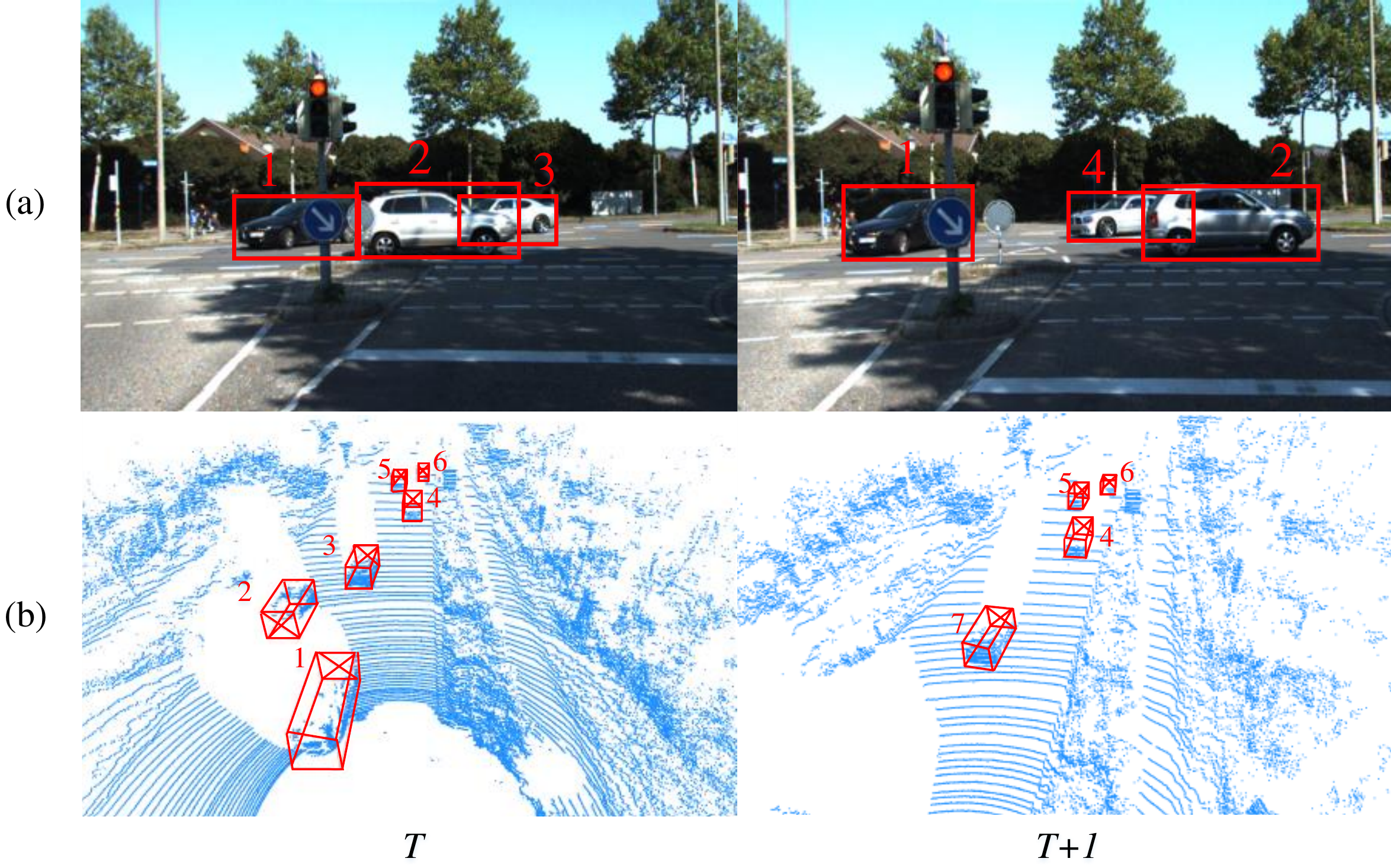}
\end{center}
   \caption{Issues encounter in 2D and 3D MOT. (a) Inaccurate appearance features can occur in camera images when objects are occluded (ID changed from 3 to 4 at the moment $T+1$). (b) The presence of large inter-frame displacements in LiDAR data can also result in tracking failures when an existing object is considered a new object (ID changed from 3 to 7 at the moment $T+1$).}
\label{fig:f1}
\end{figure}
\IEEEPARstart{3D}{multi-object} tracking (MOT) can be used to extract continuous dynamic information from surrounding environments. And it can also obtain the total number of objects in a field of view and predict the next object state to improve system reliability. Therefore, 3D MOT has been a crucial module for applications in autonomous driving~\cite{xu2021system,cao2020highway}. 

Conventional MOT algorithms first extract appearance or motion information from acquired detector results for corresponding objects. The similarity to an existing trajectory state is then calculated using an association method to achieve dynamic tracking. Camera-based MOT methods~\cite{Authors1,fang2018recurrent} are primarily applied to 2D data by using rich visual texture information to achieve stable tracking, even for irregular motion. However, feature reliability is significantly reduced when objects are occluded due to interference, as shown in Fig. \ref{fig:f1}(a). LiDAR-based MOT methods~\cite{20213D,2020Dual} are primarily adopted for 3D data and can accurately track objects, even in the presence of occlusions. However, this lack of texture information can cause mismatches during irregular object motion or large inter-frame displacements, as shown in Fig. \ref{fig:f1}(b). Therefore, camera-LiDAR fusion methods~\cite{JM3D,zhang2019robust,kim2021eagermot} are proposed to integrate the strengths of both methods. According to the fusion position, these methods can be divided into front-end and back-end fusion. Due to the difference between image and point cloud, feature alignment of the two modal data is quite difficult in front-end fusion methods~\cite{JM3D,zhang2019robust}. The back-end fusion method~\cite{kim2021eagermot} utilizes the 2D and 3D detection results from the image and LiDAR to match. However, when false detection of the detector occurs and the number of scene categories increases, tracking false detection objects and identity switches (IDS) between different categories often occur.

To address these issues, we propose a novel multi-modal MOT framework CAMO-MOT based on the combined appearance-motion optimization which not only takes advantage of information from both images and point cloud information effectively but also solves the problems of occlusion, false detection and intra-class association. CAMO-MOT consists of three main modules, including an optimal occlusion state-based object appearance module (O2S-OAM), confidence score-based motion module (CS-MM), and multi-category multi-modal fusion association module (M2-FAM). 

We design O2S-OAM to identify object occlusion states by introducing a novel occlusion head, which identifies the degree of visualization of the current object by sampling the image area of each object and feeding it into the network. By identifying the occlusion situation of the object at each moment, the appearance feature with the best occlusion situation is selected in the tracking process, thereby enhancing the robustness of the object appearance feature. The performance of the multi-object tracking algorithm depends mainly on the detector’s accuracy. Still, the existing detectors~\cite{shi2019pointrcnn,lang2019pointpillars} all have the problem of false detection, so the false detection object and the existing trajectories often generate identity switches in the multi-object tracking algorithm. Therefore, the CS-MM is proposed to reduce the influence of false detection in the motion module. \textcolor{black}{We use a robust and reasonable distance criterion——3D Generalized Intersection over Union ($gIoU_{3D}$)~\cite{Rezatofighi_2018_CVPR} to construct a motion cost matrix between detections and trajectories. Meanwhile considering the false detection object usually have minor detection confidence, we set more significant costs for low-confidence detections based on the original $gIoU_{3D}$ cost matrix to reduce the tracked possibility.} Currently, most multi-object tracking methods~\cite{hu2019joint,LGM,20213D,chaabane2021deft,huang2021joint} only provide tracking under a single category. Although a few methods \cite{kim2021eagermot,weng2020ab3dmot} provide multi-category tracking results, there are identity switches between different categories due to the scene’s complexity. To solve this problem, we propose the M2-FAM module to construct a multi-category loss, making the association only exist in the category itself. 

\textcolor{black}{Extensive experimental studies on the KITTI~\cite{Geiger2012CVPR} and nuScenes tracking benchmarks~\cite{nuscenes2019} are implemented. On the KITTI dataset, our CAMO-MOT achieves a +5.60\% HOTA and a +2.56\% MOTA compared with the famous EagerMOT~\cite{kim2021eagermot} which is the state-of-the-art multi-modal method before and ranks first among all multi-modal MOT methods. Furthermore, our method has the lowest IDS value (23 for Car and 137 for Pedestrian) compared with all other methods, which illustrates that our proposed method can maintain stable tracking in complex occlusion situations by better utilizing image and point cloud information. Furthermore, on the nuScenes dataset, our CAMO-MOT ranks \textbf{first} among all MOT methods, and achieves a +1.20\% AMOTA compared with BEVFusion~\cite{https://doi.org/10.48550/arxiv.2205.13542} which ranks second and utilizes a much more powerful detector than our detector.
Experimental results show that our CAMO-MOT performs excellently under different datasets and has strong portability. Usually, the quality of the detector has a very large impact on the performance of the tracker, however, our CAMO-MOT still achieves very competitive performance with a poor detector. We hope that CAMO-MOT can provide a simple but effective baseline algorithm for the research of tracking algorithms.}

The primary contributions of this study are as follows:
\begin{itemize}
\item We propose a novel camera-LiDAR fusion 3D MOT framework CAMO-MOT based on the combined appearance-motion optimization, which effectively integrates camera and LiDAR information to achieve stable 3D multi-object tracking for interfacing with many 3D detectors.
\item We are the first to propose an occlusion head for state estimation, which addresses the effect of occlusion in 2D images by selecting the optimal appearance feature during tracking.
\item We present a fusion association strategy that introduces a category loss to settle the interference problem for different categories in the tracking process.
\item \textcolor{black}{Our proposed CAMO-MOT algorithm has excellent performance under different tracking benchmarks. On the KITTI dataset, our method achieves state-of-the-art stability among all multi-modal MOT methods and offers the lowest IDS values among all methods. Moreover, on the nuScenes dataset, our method achieves state-of-the-art on the leaderboard among all MOT methods.}
\end{itemize}

\section{Related Work}
{\bf2D Multi-Object Tracking.} Existing image-based 2D MOT methods have benefited from the rapid development of detection algorithms~\cite{Authors1,Tang_2017_CVPR,Authors2}. Some methods~\cite{Authors2,wang2020realtime} utilize a tracking-by-detection framework, in which the tracker acquires the object region from the detector and then associates it with a trajectory using data association methods. This method often involves Kalman filters for IoUs or optical flow for matching prior to the deep learning era. These methods are simple and fast, but they fail very easily in complex scenarios.

With the advent of deep learning, many trackers have employed deep appearance features for use in associating objects. For example, DeepSORT~\cite{wojke2017simple} employs an offline trained ReID model and Kalman filters to associate objects. The Deep Affinity Network~\cite{sun2019deep} accepts two image frames as inputs, extracts appearance features under multiple perceptual object fields, and outputs a similarity score. Tracktor~\cite{Authors2} uses the Faster R-CNN as a backbone to construct a regression head for object location in the next frame, using bounding boxes from the previous frame for tracking. JDE~\cite{wang2020realtime} adopts YOLOv3~\cite{redmon2018yolov3} as a detector, adds a ReID branch to extract deep object features, and jointly trains them to improve algorithm performance. FairMOT~\cite{FairMOT} is similar to JDE in that Centernet~\cite{zhou2019objects} is utilized as a detector to further improve algorithm performance. ByteTrack~\cite{ByteTrack} leverages YOLOX~\cite{YOLOX} as a detector and a Kalman filter combined with a confidence score to perform two associations, achieving the top rank for the MOT17 dataset. However, it does not utilize deep object features in images, which causes tracking failures in complex scenes. TransTrack~\cite{TransTrack} uses a joint detection and tracking framework and the DETR~\cite{DETR} architecture as a backbone. Each object detected in the previous frame is treated as a query and is then passed to the network for use in estimating the current state. In this way, the association method is used to complete the association tracks. MOTR~\cite{MOTR} is a complete end-to-end multi-object tracking framework that utilizes video data for training. It establishes object associations in the network and considers both the appearance and loss of objects, achieving competitive results for the MOT16 dataset~\cite{MOT16}. However, the performance of these algorithms is reduced significantly when the object is occluded, due to the introduction of information outside the object itself.\\

\begin{figure*}[t]
\begin{center}
\includegraphics[width=1\textwidth]{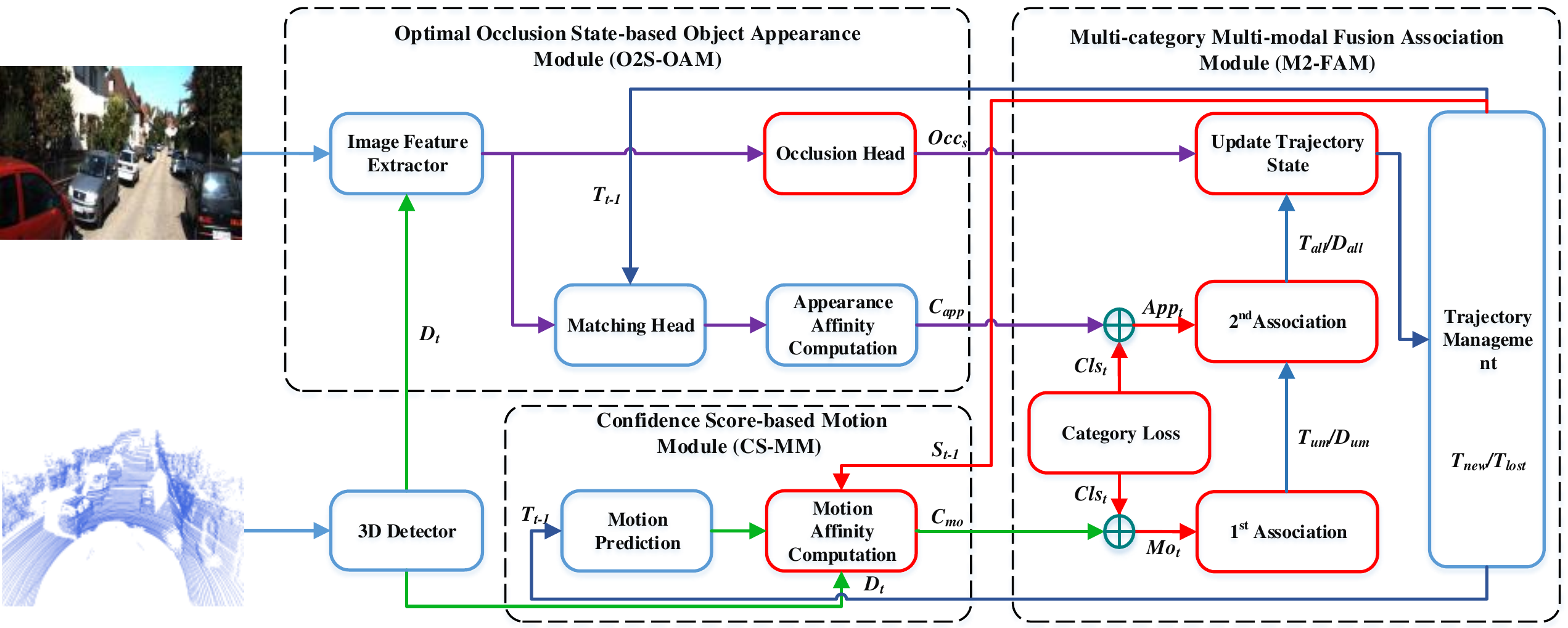}
\caption{An overview of our proposed CAMO-MOT framework in which the following steps are implemented at each discrete time $t$. (A) The 3D detector inputs the point cloud to produce the detection results $D_t$, which are then passed to O2S-OAM and CS-MM modules. (B) O2S-OAM is used to acquire deep object features from the image extractor at the frame $t$. Deep features are then passed to the occlusion head and used to estimate the occlusion state $Occ_{s}$ for each object. Following this step, the matching head outputs the appearance association matrix $C_{app}$ based on the deep features for $D_t$ and $T_{t-1}$. (C) CS-MM is used to predict the next moment state for the existing trajectory $T_{t-1}$ and solve the motion association matrix $C_{mo}$ using Euclidean distance, the existing detection $D_t$, and the confidence score $S_{t-1}$. (D) M2-FAM introduces the category loss $Cls_t$ to $C_{app}$ and $C_{mo}$ is used to obtain $App_{t}$ and $Mo_{t}$. The first association is used to output detections and trajectories based on $Mo_{t}$ and a second association is then applied to the remaining detections $D_{um}$ and trajectories $T_{um}$ based on $App_{t}$. (E) Finally, the associated trajectories $T_{all}$ are updated and the remaining detections $D_{all}$ are considered to be new trajectories.}
\label{fig:f3}
\end{center}
\end{figure*}

\indent{\bf 3D Multi-Object Tracking.} 3D MOT~\cite{scheidegger2018monocamera,wang2021joint} has a similar structure to that of 2D MOT but spatial information, which dramatically improves tracking accuracy. Motion information has been leveraged for tracking in previous studies~\cite{JRMOT,Sualeh2019Dynamic} and is achieved by associating object state estimations in a previous frame with objects in the current frame. Some methods like~\cite{JM3D} utilize an LSTM as an object state estimator to achieve tracking, but this approach often fails when objects are moving irregularly. AB3DMOT~\cite{weng2020ab3dmot} achieves excellent tracking performance using 3D Kalman filters for IoUs. Monocular cameras have also been used to estimate object distances and velocities for use as motion features in 3D MOT~\cite{Mono-Camera}. EagerMOT~\cite{kim2021eagermot} fuses 3D and 2D detection results and then applies a Kalman filter to achieve the highest HOTA on the KITTI tracking benchmark at that time, but it often leads to wrong tracking in the case of occlusion. In addition, some methods~\cite{2013Multi,2015Object} have used hand-crafted point cloud features for 3D MOT. MmMOT\cite{zhang2019robust} transfers the texture information from images into 3D spaces by fusing deep features and point clouds in different ways, to achieve various multi-modal MOT representations. JMODT\cite{huang2021joint} utilizes an attention mechanism to fuse point clouds and images, providing fused features to detectors and trackers for joint training. However, due to the sparsity of the point cloud, the information of the two modalities cannot be effectively aligned in the data processing stage, so the information of the two modalities cannot be fully utilized. PnPNet~\cite{PnPNet} uses an end-to-end 3D MOT framework to solve detection and tracking but does not utilize image information, since only point cloud data is included. However, none of these methods introduce occlusion branches to improve 3D MOT. Furthermore, most methods only consider a single tracking category, which imposes certain restrictions on tasks in complex scenes, such as cars and pedestrians being considered the same object.\\
\indent In this study, we propose a novel multi-modal MOT framework CAMO-MOT that effectively integrates image and point cloud information through two associations in which two modalities are processed independently. We are the first to introduce an occlusion head to reduce the effects of occlusions and implement multi-category loss to improve tracking effects in multi-category object scenes.
\section{CAMO-MOT}
This section presents a detailed description of the proposed 3D MOT framework CAMO-MOT, based on a tracking-by-detection framework. The algorithm consists of three primary modules, including O2S-OAM, CS-MM, and M2-FAM, as shown in Fig. \ref{fig:f3}.
\subsection{Optimal Occlusion State-based Object Appearance Module (O2S-OAM)}
\begin{figure}[t]
\begin{center}
   \includegraphics[width=1\linewidth]{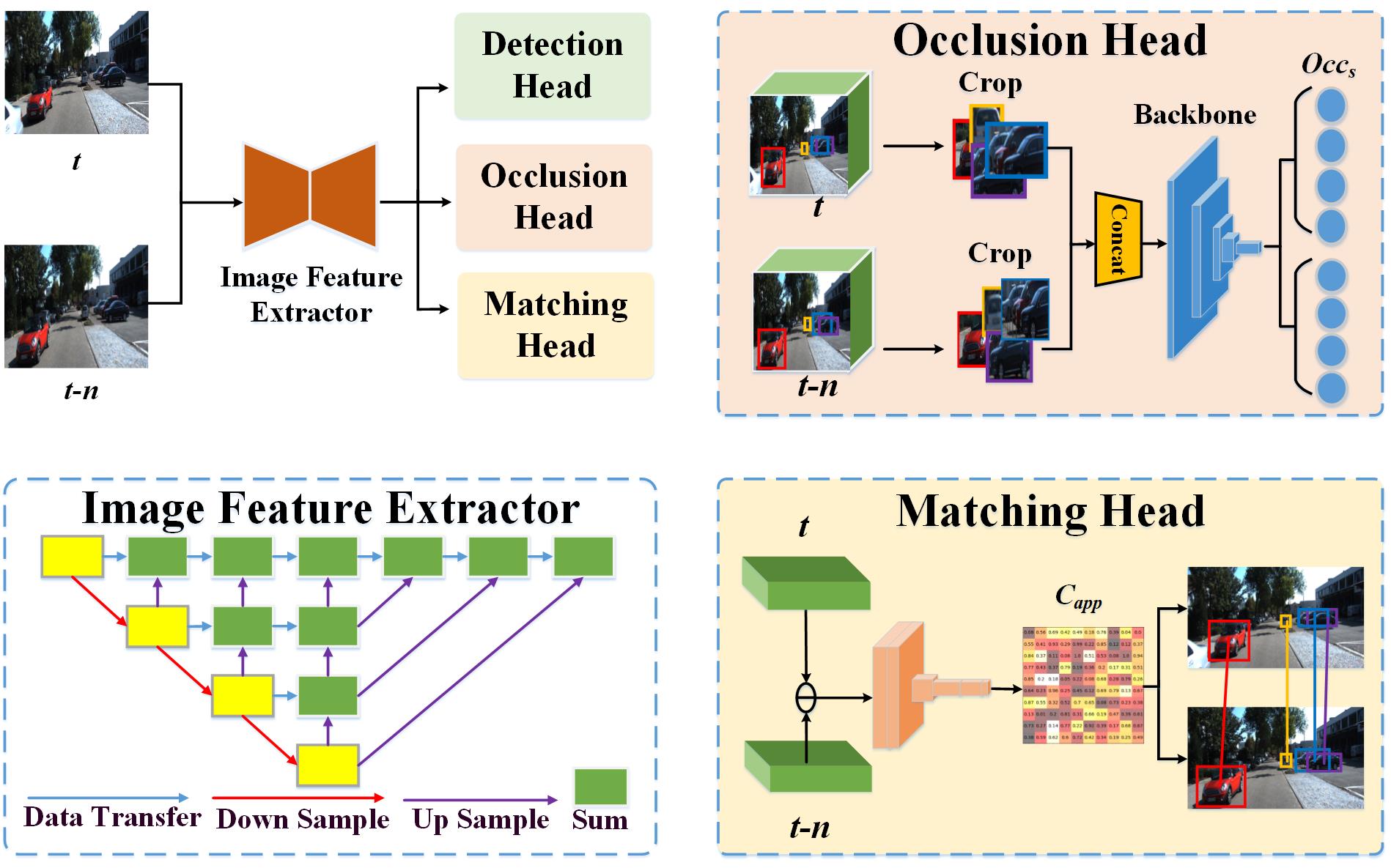}
\caption{The network structure for the optimal occlusion state-based object appearance module. The two images are used as inputs to extract deep features, which are then fed to the occlusion head as part of occlusion state recognition. The matching head is then used to associate objects in the two frames.}
\label{fig:f2}
\end{center}
\end{figure}

This section presents the details of O2S-OAM, mainly including the image feature extractor, the occlusion head, and the matching head, as shown in Fig.~\ref{fig:f2}.

1) \emph{Image Feature Extractor:} A modified DLA-34~\cite{zhou2019objects} is utilized as the image feature extractor, as shown in Fig. \ref{fig:f2}. The image at the moment $t$ is used as the input and outputs to the feature map $F_t=\{f_1,f_2,...,f_M\}$, where $M$ is the number of feature maps.

2) \emph{Occlusion Head:} Algorithm performance is typically affected when objects are occluded. This study represents the first time that an occlusion head has been introduced to identify occlusion states. In this process, a tracker is used to select optimal appearance features from multiple moments to achieve association.

Image feature maps $F_t$, acquired by the image feature extractor using the ground truth of multi-object bounding boxes $B_t=\{b_1,b_2,...,b_N\}$, are fed to the occlusion head, where $N$ is the number of objects. This process is shown in Fig. \ref{fig:f2}. In order to improve the accuracy of the algorithm, the image area of the object is then bilinearly interpolated to $224\times224$ and sent to the backbone to identify occlusion states. The term $Occ_{s}$ is then generated for each object and 
used to determine its occlusion state, which is used in online tracking for optimal appearance feature selection.\\
3) \emph{Matching Head:} Most existing association methods calculate similarity values by extracting ReID features used to associate objects. However, these methods do not sufficiently consider the relationships between objects. As such, we utilize an end-to-end association network to directly output an association matrix, which dramatically improves algorithm generality, as shown in Fig.~\ref{fig:f2}. This matching head is similar to that of prior studies\cite{sun2019deep,chaabane2021deft} and uses object feature embedding in the association of detection results between two frames, but the distinction is that we adopt the difference between features of the two frames as the input as the difference can better represent the degree of correlation of features. These embedded data are then extracted using the feature map acquired by the image feature extractor. The position of the object center in an original image of input size $W \times H$ is given by $(x,y)$, which is then mapped to the $m^{th}$ feature map of size $W_m\times H_m\times C_m$. This position is then shifted to $(\frac{x}{W}W_m,\frac{y}{H}H_m)$ during extraction of the $C_m$-dimensional vector $o_m$.

Since the number of objects in each frame is not unique, $N_{max}$ is set as the maximum number of detections per frame. The tensor $E_{t,t-n}\in{ R^{(C\times M)\times N_{max}\times N_{max}}}$ is reconstructed by concatenating the differences between object features in frames $t$ and $t-n$, and using a zero tensor to fill in any numbers less than $N$. A $1\times1$ convolutional neural network is adopted to produce the final association cost matrix $A_{t,t-n}\in{R^{N_{max}\times N_{max}}}$. A row and a column are then added to $A_{t,t-n}$ and the parameters that the network can learn is used to supplement the matrix with objects that are not associated (i.e., new objects entering or old objects leaving a scene~\cite{sun2019deep,chaabane2021deft}). A $Softmax$ function is then used to process each row and column separately, to produce the final association matrices $\hat{A}_t,\hat{A}_{t-1}\in{R^{(N_{max}+1)\times (N_{max}+1)}}$. The averages of these matrices are used as the final affinity scores.
\subsection{Confidence Score-based Motion Module (CS-MM)}
Conventional MOT algorithms often do not consider false detection problems, which typically involve low confidence scores. In this study, a confidence score is used to jointly solve a cost matrix. Objects with lower confidence scores have higher costs and are more challenging to track continuously. The included Kalman filters are optimized to obtain a prediction score for the trajectory state. The resulting motion prediction can be expressed as:
\begin{equation}
\hat{Z}_{t}^{j} = A{Z}_{t-1}^{j},
\end{equation}
\begin{equation}
\hat{S}_{t}^{j} = A{S}_{t-1}^{j}A^T+Q,
\end{equation}
\begin{equation}
A=
\begin{bmatrix}
  E_{3\times3} & \sigma E_{3\times3} & \frac{1}{2}\sigma^2 E_{3\times3} &   \\
  O_{3\times3} & E_{3\times3} & \sigma E_{3\times3} & O_{3\times n} \\
  O_{3\times3} & O_{3\times3} & E_{3\times3} &   \\
    & O_{n\times9} &   & E_{n\times n}
\end{bmatrix},
\end{equation}
where $\hat{Z}_{t}^{j}$ is the estimated state for the current moment, ${Z}_{t-1}^{j}$ is the previous state of each tracked result, $A$ is the state transformation matrix, ${S}_{t-1}^{j}$ is the corresponding error covariance, $\hat{S}_{t}^{j}$ is the error covariance estimated for the current moment, $Q$ is the covariance matrix for the state function, $E$ and $O$ represent the unit and zero matrices, \textcolor{black}{$\sigma$ is the interval time between two adjacent frames of the LiDAR scan, and $n=4+k$, 4 represents the other tracking information (e.g. length $l$, width $w$, height $h$, and rotation angle $\alpha$), $k$ represents position ($x, y, z$), velocity ($vx, vy, vz$), and acceleration ($ax, ay, az$) of an object in 3D space.}

Generally speaking, an object with a low confidence score has a much lower probability of being associated. As such, the motion cost matrix is divided by the state score, in which trajectories with a lower state score exhibit higher losses when solving for optimal association pairs. The corresponding motion cost matrix can be calculated as follows:

\begin{equation}
\begin{aligned}
    M_t^{i,j} &= 1 - gIoU_{3D}(D_{t}^{i},\hat{T}_{t}^{j}) \\
              &= 2 - \frac{V(D_{t}^{i} \cap  \hat{T}_{t}^{j})}{V(D_{t}^{i} \cup \hat{T}_{t}^{j})} - \frac{V(D_{t}^{i} \cup  \hat{T}_{t}^{j})}{V_{hull}(D_{t}^{i},\hat{T}_{t}^{j})} \\
  \end{aligned}
\end{equation}
\begin{equation}
C_{mo}=
\begin{bmatrix}
  M_t^{1,1}/\gamma_t^1 & \cdots & M_{t}^{1,N_{t-1}^Y}/\gamma_{t}^{N_{t-1}^Y} \\
  \vdots & \ddots & \vdots \\
   M_t^{N_t^X,1}/\gamma_t^1 & \cdots &  M_t^{N_t^X,N_{t-1}^Y}/\gamma_t^{N_{t-1}^Y}
\end{bmatrix},
\end{equation}
\begin{equation}
\gamma_t^\sigma=\{
\begin{array}{ccc}
  1, & c^{\sigma}_{t-1}=0 & c_t^\sigma\ne0 \\
  \gamma_t^\sigma+\theta c^{\sigma}_{t} &   & otherwise\\
\end{array},
\end{equation}
where $M_t$ is the cost matrix of the motion calculated with $gIoU_{3D}$ at moment $t$. $D_{t}^{i}=\left \{ x, y, z, w, h, l, \alpha, (vx, vy) \right \}$ is the $i$-th 3D detection bounding box containing the center position $(x, y, z)$, and the 3D size(width, length, height) $(w, h, l)$, and the rotation angle $\alpha$, and the velocity $(vx, vy)$ on the ground plane of the current frame $t$, note that whether velocity information is included or not depends on the dataset. $\hat{T}_{t}^{j}$ is the estimated state containing information described in the previous paragraph of the $j$-th trajectory predicted by Kalman filtering for the current frame $t$. $V(D_{t}^{i} \cap  \hat{T}_{t}^{j})$, $V(D_{t}^{i} \cup \hat{T}_{t}^{j})$ are the intersection and union volumes of $D_{t}^{i}$ and $\hat{T}_{t}^{j}$, respectively. $V_{hull}(D_{t}^{i}$ is the volume of the convex hull computed by $D_{t}^{i}$ and $\hat{T}_{t}^{j}$. $N_t^X$ is the number of objects detected in the current frame. $N_{t-1}^Y$ is the number of trajectories at present. $c_t^\sigma \in [0,1]$ is the detection confidence score. $\gamma_t^\sigma$ is the state prediction score for the current trajectory (representing the degree of reliability). $\theta$ is the trajectory state decay factor. $C_{mo}$ is the final output motion cost matrix of CS-MM.

\subsection{Multi-category Multi-modal Fusion Association Module (M2-FAM)}
A 3D motion cost matrix is first applied to estimate the position of the object in 3D space to compensate for the effects of occlusions in 2D space by predicting the 3D space position of the object at the last moment. An appearance cost matrix is then used to solve tracking failures caused by large inter-frame displacements and irregular motion of the object in 3D space. Interference between categories often leads to tracking failures in MOT tasks, particularly for objects that are densely distributed with multiple categories. To address these issues, we first introduce a category loss function so that objects are associated only within categories that improve algorithm robustness in complex multi-category scenes.

In this process, individual categories, are assigned different constants (e.g. 0, 1, 2) as their category indicators. It is important to note that a large gap (e.g. $10^5$) is included between the constants to ensure no inter-class interference is present in subsequent associations. The resulting association framework is shown in Fig. \ref{fig:f3}. The motion cost matrix $C_{mo}$ and the appearance cost matrix $C_{app}$ are acquired from CS-MM and O2S-OAM, respectively. Category loss is then separately added to the outputs $Mo_t$ and $App_t$ for subsequent associations, calculated using the following equations:
\begin{equation}
Dis_t^{i,j}= D_{t}^{j}(cls)-T_{t-1}^{j}(cls),
\end{equation}
\begin{equation}
Cls_t^{i,j} = 
\begin{cases}
  10^5, & Dis_t^{i,j}\ne0 \\
  0 & otherwise\\
\end{cases},
\end{equation}
\begin{equation}
Mo_t=C_{mo}+Cls_t,
\end{equation}
\begin{equation}
App_t=C_{app}+Cls_t,
\end{equation}
\textcolor{black}{where $Dis_t$ is the difference between classes corresponding to the detections and the all active trajectories at moment $t$}, $cls$ is the index of the corresponding category, and $Cls_t$ is category loss.

The $Mo_t$ and $App_t$ terms are acquired in the association phase. We first associate trajectories based on $Mo_t$ to acquire the corresponding detections. We then set the existing trajectory as $T_{t-1}$, the existing detection as $D_t$, and use $App_t$ to associate the remaining detections $D_{um}$ and trajectories $T_{um}$. This process can be represented as:
\begin{equation}
\begin{aligned}
&\min\,\,  \sum_{i=1}^{M}\sum_{j=1}^{N}Mo_t^{i,j}+\sum_{i=1}^{M'}\sum_{j=1}^{N'}App_t^{i,j},\\
&s.t.\quad
\begin{cases}
   Mo_t^{i,j} \leq \theta_{mo}\\
   App_t^{i,j} \leq \theta_{App},
\end{cases}
\end{aligned}
\end{equation}
where $N$ and $M$ are the number of detections and trajectories at the moment $t$, respectively, $N'$ is the number of unassociated detections, $M'$ is the number of unassociated trajectories, and $\theta_{mo}$ and $\theta_{App}$ are the maximum association costs for the motion and appearance modules, respectively.

Appearance features and occlusion states are maintained for each moment of the tracking process because occlusion features that contain information other than the object itself are not reliable. When updating the trajectory, we select the optimal appearance features among a lower occlusion state among multiple trajectory moments. These are assumed to be the final trajectory features to further improve algorithm robustness by eliminating the effects of occlusions.

\subsection{Training}
During training, pairs of images with an interval of n frames are used as inputs to the O2S-OAM, as shown in Fig. \ref{fig:f2}. This interval between image pairs is set as a random number of frames ($1\le n \le n_{max}$) to promote robustness to temporary occlusions or missed detections~\cite{chaabane2021endtoend,chaabane2021deft}.
ResNet18 is adopted as the backbone for the occlusion head, with cross-entropy serving as its loss function $\emph{\L}_{occlusion}$. The specific form is as follows:
\begin{equation}
\emph{\L}_{occlusion}=-\frac{1}{N}\sum_{i=1}^{N}\sum_{c=1}^{4}y_{ic}log(p_{ic}),
\end{equation}
where $y$ is the label and $p$ is the probability vector output by the model.

In each training phase, we generate a ground truth association matrix $M_g$ of size $(N_{max}+1)\times(N_{max}+1)$, consisting of entries $[i,j]\in {0,1}$. Values of 1 and 0 in the matrix indicate the object is associated and  when $i$ and $j$ are less than $N_{max}$, respectively. A value of 1 in the matrix (for $i$ or $j$ equal to $N_{max}+1$) indicates the object is lost or newly emerged and should not be associated.

Two frames are input and used to denote $\emph{\L}_{match_t}$ and $\emph{\L}_{match_{t-n}}$ as the associated matrices in frames $t$ and $t-n$, respectively. Considering both should be symmetric, we apply an average as the final association loss function $\emph{\L}_{match}$:
\begin{equation}
\emph{\L}_{match}= \frac{\emph{\L}_{match_t}+\emph{\L}_{match_{t-n}}}{2(N_t+N_{t-n})},
\end{equation}
\begin{equation}
\emph{\L}_{match_t}=\sum_{i=1}^{N_{max}}\sum_{j=1}^{N_{max}+1}M_g(i,j)log(Softmax(\hat{A}(i,j))),
\end{equation}
\begin{equation}
\emph{\L}_{match_{t-n}}=\sum_{i=1}^{N_{max}+1}\sum_{j=1}^{N_{max}}M_g(i,j)log(Softmax(\hat{A}(i,j))),
\end{equation}
where $N_t$ and $N_{t-n}$ represent the number of objects at time $t$ and time $t-n$, respectively. The final loss function $\emph{\L}_{join}$ can then be expressed as follows:
\begin{equation}
\emph{\L}_{join}=\frac{1}{e^{\lambda_1}}\emph{\L}_{detect}+\frac{1}{e^{\lambda_2}}(\emph{\L}_{match}+\emph{\L}_{occlusion})+\lambda_1+\lambda_2,
\end{equation}
where $\emph{\L}_{detect}$ is the CenterNet detection head loss and $\lambda_1$ and $\lambda_2$ are the learnable loss weights for detection and other branching tasks (e.g. occlusion head and matching head).

\section{Experiments}

\textcolor{black}{In order to fully verify the performance of our proposed CAMO-MOT algorithm, we evaluate it on the famous KITTI and nuScenes tracking datasets: (i) KITTI 2D MOT, (ii) KITTI 3D MOT, (iii) nuScenes 3D MOT. First, we introduce the two datasets and the evaluation metrics. Then, the implementation details of our method on each dataset are described. Finally, comparative experiments with state-of-the-art methods on the benchmarks and adequate ablation studies are presented. In addition, we also provide qualitative visualizations to illustrate the effectiveness.
} 

{\bf KITTI.} \textcolor{black}{
KITTI tracking benchmark~\cite{Geiger2012CVPR} provides 21 training sequences and 29 test sequences, in which each sequence consists of hundreds of frames.}
The KITTI training set contains 8,008 frames and the test set includes 11,095 frames. The input used in these experiments includes images, point clouds, and IMU/GPS data. \textcolor{black}{KITTI provides the ground truth of the training set at a frequency of 10 Hz. For 2D MOT tracking benchmark, the result of the algorithm in the test set needs to be submitted to the KITTI official website, and HOTA~\cite{HOTA} is the primary metric. For 3D MOT tracking benchmark, we follow the evaluation protocol of~\cite{weng2020ab3dmot}, using AMOTA~\cite{weng2020ab3dmot} as the primary metric.}

Training sequences 1, 6, 8, 10, 12, 13, 14, 15, 16, 18, and 19 are applied as the validation set~\cite{kim2021eagermot,weng2020ab3dmot}, while other sequences are regarded as the training set. Our algorithm is applied to analyze cars and pedestrians, in agreement with the KITTI test set evaluation.

\textcolor{black}{{\bf nuScenes.} 
The nuScenes~\cite{nuscenes2019} tracking benchmark consists of 850 training sequences and 150 test sequences. 
Each sequence consists of 40 frames approximately (2Hz in 20 seconds), which further assesses the robustness of the MOT algorithm under various scenarios. The training set contains 34,149 frames, while the test set includes 6,008 frames. Nuscenes samples keyframes (image, LiDAR, RaDAR) at a rate of 2Hz and delivers annotation information for each keyframe. The keyframe frequency of 2Hz is not favorable to the precise prediction of the motion model and introduces a significant inter-frame displacement, posing a formidable design problem for the 3D MOT algorithm. For 3D MOT Tracking benchmark, the algorithm's performance on the test set must be evaluated using the official evaluator, which uses AMOTA~\cite{weng2020ab3dmot} as the primary evaluation metric.}

\textcolor{black}{In contrast to KITTI, we segregate the validation set from the training set using the official script\footnote{https://github.com/nutonomy/nuscenes-devkit/blob/master/python-sdk/nuscenes/utils/splits.py}. The validation set includes 150 scenes, 6,019 frames, and 140k instances of object annotations. Our algorithm tracks seven categories specified by nuScenes.}

{\bf Evaluation Metrics.} \textcolor{black}{For KITTI 2D tracking benchmark, we mainly follow the rules of CLEAR MOT~\cite{bernardin2008evaluating} and HOTA~\cite{HOTA} for evaluation. The Higher-Order Tracking Accuracy (HOTA) comprehensively evaluates the performance of the tracker considering the impact of different detection thresholds on tracking.} The HOTA is defined as:
\begin{equation}
\left\{
\begin{aligned}
  &TPA(c) = \{k\},\\
  &k\in\{TP\mid prID(k)=prID(c)\wedge gtID(k)=gtID(c)\},\\
  &FNA(c) = \{k\},\\
  &k\in
  \begin{array}{l}
  \{TP\mid prID(k)\ne prID(c)\wedge gtID(k)=gtID(c)\}\\
  \bigcup\{FN\mid gtID(k)=gtID(c)\},
  \end{array}\\
  &FPA(c) = \{k\},\\
  &k\in
  \begin{array}{l}
  \{TP\mid prID(k)=prID(c)\wedge gtID(k)\ne gtID(c)\}\\
  \bigcup\{\textcolor{black}{FP}\mid prID(k)=prID(c)\},
  \end{array}
\end{aligned}
\right.
\end{equation}
\begin{equation}
HOTA = \sqrt{\frac{\sum_{c\in{TP}}A(c)}{\mid TP\mid+\mid FN\mid+\mid FP\mid}},
\end{equation}
\begin{equation}
A(c) = \frac{\mid TPA(c)\mid}{\mid TPA(c)\mid+\mid FNA(c)\mid+\mid FPA(c)\mid},
\end{equation}
where TP means true positives, FN means false negatives, FP means false positives, TPA means true positive associations, FNA means false negative associations, FPA means false positive associations, prID is the predicted identity, and gtID is the actual identity. 

\textcolor{black}{CLEAR-MOT consists mostly of two exhaustive evaluation metrics: MOTA and MOTP.} The Multi-Object Tracking Accuracy (MOTA) measures the overall tracking accuracy and is defined as:

\begin{equation}
MOTA = 1-\frac{\sum_{t}(F_t+N_t+IDs_t)}{\sum_{t}G_t},
\end{equation}
where $F_t$, $N_t$, \textcolor{black}{$IDs_t$ and $G_t$} represent false positives, the number of misses, mismatches and ground truth, respectively. 

The value of Multi-Object Tracking Precision (MOTP) is measured by the overlapping ratio between the estimated object and its ground truth, and is defined as:
\begin{equation}
MOTP = \frac{\sum_{t}^{i}c_t^i}{\sum_{t}M_t},
\end{equation}
where $c_t^i$ represents the distance between detection and its corresponding ground truth, and $M_t$ is the number of all matches. Additionally, we employ five more metrics, MT (mostly tracked), ML (mostly lost), FP (false positives), FN (false negatives), and IDS (identity switches), aiming for a better comparison of the tracker performance.

\textcolor{black}{For KITTI 3D MOT and nuScenes 3D MOT, we mainly follow the rules of averaged MOTA and MOTP (AMOTA and AMOTP~\cite{weng2020ab3dmot}), which can completely evaluate the overall accuracy and robustness of the tracker under various recall thresholds. AMOTA is defined as:}

\begin{equation}
AMOTA = \frac{1}{L} \sum_{r \in \left \{ \frac{1}{L}, \frac{2}{L}, \cdots , 1   \right \} }MOTA_{r},    
\end{equation}
\textcolor{black}{where $MOTA_r$ represents the MOTA computed at a specific recall value $r$. Since $MOTA_r$ has a strict upper bound of $r$, in order to get metrics ranging from 0\% to 100\%, $MOTA_r$ and AMOTA need to be scaled numerically. $sMOTA_r$ (scaled MOTA) and sAMOTA (scaled AMOTA) are defined as follows:}

\begin{equation}
sMOTA_r = max(0, \frac{MOTA_r}{r}).
\end{equation}
\begin{equation}
sAMOTA = \frac{1}{L} \sum_{r \in \left \{ \frac{1}{L}, \frac{2}{L}, \cdots , 1   \right \} }sMOTA_{r}.
\end{equation}
\textcolor{black}{Note that the AMOTA in the official nuScenes tracking benchmark is in fact the sAMOTA in~\cite{weng2020ab3dmot}.}

\subsection{Implementation Details}

\textcolor{black}{Our tracking method is implemented in Pytorch for all benchmarks and it is trained and evaluated on an Ubuntu server equipped with RTX 3090 GPU. During training, 13 feature maps are extracted from the modified DLA-34 embedded backbone~\cite{chaabane2021deft}. We also transport the final feature map to the occlusion head in order to recognize the occlusion state. The data are augmented using random cropping, flipping, scaling, and photometric distortions to increase robustness.}

{\bf KITTI.} During training, CAMO-MOT is trained for 100 epochs using the Adam~\cite{kingma2017adam} optimizer with an initial learning rate of 1.25$e^{-4}$ and a batch size of 8. At 60 epochs, the learning rate is decreased by $e^{-5}$.

\textcolor{black}{During inference, our method can run at ~25 FPS per image with an RTX 3090 GPU. The selection of hyperparameters is based on the highest HOTA scores found in the validation set. 3D detection results are filtered using Non-Maximum Suppression (NMS) with a threshold of 0.1 ($\theta_{nms} = 0.1$). We use $\theta_{mo} = 0.01$, and $\theta_{App} = 1.4$ as thresholds for the Hungarian algorithm to match motion and appearance matrices. A new trajectory will be created for the unmatched detection in the current frame. A trajectory is discarded if it has not been updated in the last 15 frames. For each trajectory's unmatched frames, the motion model's predicted trajectory state is appended to the result file.}

\textcolor{black}{{\bf nuScenes.} During training, we adopt the same optimizer (Adam~\cite{kingma2017adam}) and batch size (8) as on KITTI dataset. We train our method with the learning rate of 1.25$e^{-4}$ for 60 epochs.}

\textcolor{black}{
Hyperparameters are chosen based on the best AMOTA score identified in the validation set. We evaluate CAMO-MOT with a variety of 3D detectors. Confidence Score Filter (CSF) and NMS are applied to the input 3D detection boxes. Each object's 3D bounding box is projected onto all six cameras, and the 2D bounding box with the biggest projected area is used as the object's image feature. We use $\theta_{nms} = 0.08$ for all 3D detectors. The threshold for confidence score filter is detector-specific:  $\theta_{CSF} = [0.03, 0.12, 0.14]$ for BEVFusion~\cite{https://doi.org/10.48550/arxiv.2205.13542}, FocalsConv~\cite{focalsconv-chen}, CenterPoint~\cite{yin2021center}. We use $\theta_{mo} = 0.02$, and $\theta_{App} = 1.4$ for all seven classes. Trajectory initialization is consistent with the scheme on the KITTI dataset. We adopt a mixed scheme of count-based and confidence-based~\cite{benbarka2021score} to discard trajectories, which means that a trajectory will be discarded when it has not been updated in the last 15 frames or the tracking score falls down the deletion-threshold ($\theta_{del} = 0$). For matching frames in the trajectory, we output the motion model's updated trajectory state. We complete the trajectory state for each mismatch period and output up to 2 frames of the trajectory state predicted by the motion model during each mismatch period. Similar to SimpleTrack~\cite{pang2021simpletrack}, the predicted state's tracking score is equal to 0.05 times the score of the most recently updated trajectory state. To reduce false positives, we apply NMS with $\theta_{nms} = 0.08$ to all output trajectory states.}

\subsection{Comparative Evaluation}

\textcolor{black}{\textbf{KITTI 2D MOT.}} A series of experiments are implemented on the KITTI tracking benchmark test set to evaluate our proposed CAMO-MOT algorithm, the results of which are provided in Tables~\ref{table:1a} and~\ref{table:2a}. As shown, our method ranks first place and surpasses state-of-the-art tracking methods with remarkable margins on the KITTI tracking benchmark test set.

\begin{table*}[t]
\caption[l]{
{A comparison of existing algorithms applied to the KITTI tracking benchmark test set (Car class). Methods utilizing the same detector (PointGNN) are labeled by ``*", while ``Delta" provides a comparison between our method and EagerMOT~\cite{kim2021eagermot}.}}
\label{table:1a}
\setlength\tabcolsep{9pt}
\begin{tabular}{lllllllllll}
\hline
\bf{Method}     & \bf{Modality}     & \bf{Type} & \bf{HOTA$\uparrow$}    & \bf{MOTA$\uparrow$}      & \bf{MOTP$\uparrow$}     & \bf{MT$\uparrow$}    & \bf{FN$\downarrow$}   & \bf{IDS$\downarrow$}   &\bf{ML$\downarrow$} &\bf{FP$\downarrow$}  \\
\hline
AB3DMOT~\cite{weng2020ab3dmot}    & LiDAR        & 3D   & 69.81 & 83.49    & 85.17 & 67.08   & 1060 & 126 &11.38 &4492  \\
JMODT~\cite{huang2021joint}      & LiDAR\&Camera & 3D   & 70.73 & 85.35    & 85.37 & 77.39   & 1249 & 350 &2.92 &3438\\
LGM~\cite{LGM}  & Camera & 2D   & 73.14  & 87.60      & 84.12  & 85.08    & 1568  & 448 &2.46   &\bf{2249} \\
mono3DT~\cite{hu2019joint}      & Camera & 3D   & 73.16 & 84.28    & 85.45 & 73.08   & 745 & 379 &2.92 &4282\\
DEFT~\cite{chaabane2021deft}       & Camera       & 2D   & 74.23 & 88.38    & 84.46 & 84.31   & 1006 & 344 &\bf{2.15} &2647 \\
EagerMOT*~\cite{kim2021eagermot}  & LiDAR\&Camera & 3D   & 74.39 & 87.82    & \bf{85.69} & 76.15   & 454  & 239 &2.46 &3497 \\
Mono\_3D\_KF~\cite{Mono_3D_KF}  & Camera & 3D   & 75.47 & 88.48    & 83.70 & 80.61   & 1045  & 162 &4.15  &2754 \\
OC-SORT~\cite{OC-SORT}       & Camera        & 2D   &76.54 & 90.28    & 85.53 & 80.00   & 407  & 250 &3.08 & 2685\\
PC3T~\cite{20213D}       & LiDAR        & 3D   & 77.80 & 88.81    & 84.26 & 80.00   & 814  & 225 &8.46 &2810  \\
PermaTrack~\cite{Permanence}       & Camera        & 2D   & 78.03 & \bf{91.33}    & 85.65 & \bf{85.69}   & \bf{402}  & 258 &2.62 &2320
\\\hline
Ours*      & LiDAR\&Camera & 3D   & \bf{79.95} & 90.38    & 85.00 & 84.61  & 962  & \bf{23}  &7.38 &2322  \\
\rowcolor{mypink} Delta & LiDAR\&Camera & 3D   & +5.56 & +2.56    & -0.69 & +8.46  & +508  & -216  &+4.92 &-1175  \\
\hline
\end{tabular}
\end{table*}

\begin{table*}[h]
\caption{{}{A comparison of existing algorithms on the KITTI tracking benchmark test set (Pedestrian class). Methods using the same detector (PointGNN) are labeled by ``*", and ``Delta" is the comparison between our method and EagerMOT~\cite{kim2021eagermot}.}}
\label{table:2a}
\setlength\tabcolsep{9pt}
\begin{tabular}{lllllllllll}
\hline
\bf{Method}      & \bf{Modality}     & \bf{Type} & \bf{HOTA$\uparrow$}        & \bf{MOTA$\uparrow$}    & \bf{MOTP$\uparrow$}       & \bf{MT$\uparrow$}      & \bf{FN$\downarrow$}    & \bf{IDS$\downarrow$}  &\bf{ML$\downarrow$} &\bf{FP$\downarrow$} \\
\hline
AB3DMOT~\cite{weng2020ab3dmot}      & LiDAR\&Camera        & 3D   & 35.57  & 38.93    & 64.55    & 17.18   & 2135 & 259  &41.24 &11744 \\
EagerMOT*~\cite{kim2021eagermot}    & LiDAR\&Camera & 3D   & 39.38  & 49.82    & 64.42    & 27.49   & 2161 & 496 &24.05 &8959 \\
TrackMPNN~\cite{TrackMPNN}     & Camera & 2D   & 39.40   & 52.10     & 73.42     & 35.05    & 2758 & 626 &18.90  &7705\\
JCSTD~\cite{JCSTD}  & Camera       & 2D   & 39.44  & 43.42     & 71.72    & 18.56   & \bf{885} & 236   &34.36 &11976\\
CenterTrack~\cite{2020Tracking}  & Camera       & 2D   & 40.35  & 53.84    & 73.72   & 35.40   & 2201 & 425   &21.31 &8061\\
QD-3DT~\cite{hu2021monocular}  & Camera       & 3D   & 41.08  & 51.77    & 73.13    & 32.65   & 2084 & 717   &19.24 &8364\\
Quasi-Dense~\cite{pang2021quasidense}  & Camera       & 2D   & 41.12  & \bf{55.55}    & \bf{73.70}    & 31.27   & 1309 & 487   &19.24 &8364\\
MDP~\cite{2015Learning}  & Camera       & 2D   & 42.76  & 47.02    & 70.17    & 25.77   & 2502 & 213   &28.52  &9550\\
Mono\_3D\_KF	~\cite{Mono_3D_KF}  & Camera       & 3D   & 42.87  & 45.44    & 69.06     & 33.68   & 3498 & 267   &26.46  &8865\\
NC2~\cite{NC2}     & LiDAR & 3D   & 44.30  & 44.18    & 65.68    & \bf{44.33}   & 6415 & 332 &\bf{13.06} &6176
\\\hline
Ours*       & LiDAR\&Camera & 3D   & \bf{44.90}  & 52.19    & 64.50    & 35.40  & 2560 & \bf{137}  &25.43 &\bf{1112} \\
\rowcolor{mypink} Delta & LiDAR\&Camera & 3D   & +5.52 & +2.37    & +0.08 & +7.91  & +399  & -359  &+1.38 &-7847  \\
\hline
\end{tabular}
\end{table*}

Our proposed method compares with EagerMOT~\cite{kim2021eagermot} which is the state-of-the-art multi-modal algorithm currently and we adopt the identical detector to ensure competitive balance. Our method achieves a HOTA of 79.95\%, a MOTA of 90.38\%, and an IDS of 23 for the Car class, as shown in Table~\ref{table:1a}. It also achieves a HOTA of 44.90\%, a MOTA of 52.19\%, and an IDS of 137 for the Pedestrian class, as shown in Table~\ref{table:2a}. These represent HOTA increases of +5.56\% and +5.52\% and MOTA increases of +2.56\% and +2.37\% for the Car and Pedestrian classes, respectively, compared with EagerMOT~\cite{kim2021eagermot}. These results indicate our method offers strong tracking capabilities, achieving the lowest IDS values among all methods. This suggests we can achieve stable tracking with less mismatching, dramatically improving the reliability.

\textcolor{black}{\textbf{KITTI 3D MOT.} Following the evaluation protocol by~\cite{weng2020ab3dmot},} we provide the 3D MOT results of our method on the KITTI dataset. As very few methods give the 3D MOT results, we compare with AB3DMOT~\cite{weng2020ab3dmot}, as shown in Table~\ref{table:5a}. It is evident our method achieves sAMOTA increases of +2.01\% for the Car class and +13.98\% for the Pedestrian class, illustrating the remarkable performance of our method.

\begin{table}[t]
\caption{3D MOT evaluation on the KITTI validation set. ``Delta" is the comparison between our method and AB3DMOT~\cite{weng2020ab3dmot}. (AMOTA (average multi-object tracking accuracy), AMOTP (average multi-object tracking precision), sAMOTA (scaled AMOTA)}
\label{table:5a}
\setlength\tabcolsep{8pt}
\begin{tabular}{lllll}
\hline
\bf{Method}   &\bf{Type}    & \bf{sAMOTA$\uparrow$} & \bf{AMOTA$\uparrow$}  & \bf{AMOTP$\uparrow$}  \\ \hline
AB3DMOT~\cite{weng2020ab3dmot}  &\multirow{3}{*}{Car}& 93.28  & 45.43 & 77.41       \\
Ours     && \bf{95.29}       &\bf{48.04}        &\bf{81.48}  \\
\rowcolor{mypink}  Delta   && +2.01       &+2.61        &+4.07    \\ \hline
AB3DMOT~\cite{weng2020ab3dmot}  &\multirow{3}{*}{Ped.}& 75.85  & 31.04 & 55.53    \\
Ours     &&\bf{89.83}       &\bf{44.84}       &\bf{72.55}   \\
\rowcolor{mypink}  Delta   && +13.98       &+13.80        &+17.02    \\ \hline
\end{tabular}
\end{table}

\begin{table}[t]
\caption{{}
{A universality assessment for our proposed method applied to the validation set in the KITTI tracking benchmark. }}
\label{table:3a}
\setlength\tabcolsep{5pt}
\begin{tabular}{lllllll}
\hline
\bf{Detector}        & \bf{Type}       & \bf{HOTA$\uparrow$}  & \bf{MOTA$\uparrow$}  & \bf{IDS$\downarrow$}   &\bf{FP$\downarrow$}  &\bf{FN$\downarrow$}\\ \hline
SECOND~\cite{lang2019pointpillars}    &  \multirow{4}{*}{Car}         & 75.71    & 80.02   &   12 &\bf{237}&1425 \\
PVRCNN~\cite{shi2021pvrcnn}    &         & 79.27    & 87.03   &   10 &589&588 \\
PointRCNN~\cite{shi2019pointrcnn} &         & 77.99    & 86.31   &   9   &680&558\\
PointGNN~\cite{shi2020pointgnn}  &         & \bf{80.74}  & \bf{87.78}   &   \bf{6} &544&\bf{474}  \\ \hline
SECOND~\cite{lang2019pointpillars}   & \multirow{4}{*}{Ped.} & 26.19    & 17.31   &   \bf{8} &\bf{130}&7955  \\
PVRCNN~\cite{shi2021pvrcnn}   &  & 31.81    & 23.49   &   13 &331&7144 \\
PointRCNN~\cite{shi2019pointrcnn} &    & 45.65    & 61.54   &   95 &887&2792  \\
PointGNN~\cite{shi2020pointgnn}        &  & \bf{50.24}    & \bf{57.19}   &   88 &2850&\bf{1256} \\ \hline
\end{tabular}
\end{table}

\textcolor{black}{\textbf{nuScenes 3D MOT}. We also evaluate our CAMO-MOT method on the nuScenes tracking benchmark test set. The results are shown in Table~\ref{table:10a}, in which most state-of-the-art MOT methods are given. Our CAMO-MOT achieves the AMOTA of 75.3\% and ranks first among all algorithms on the nuScenes tracking benchmark test set. On the test set, we fuse BEVFusion~\cite{https://doi.org/10.48550/arxiv.2205.13542} and FocalsConv~\cite{focalsconv-chen} according to the detection class and use the result as the 3D detector. Despite employing a lower detector than BEVFusion, we achieve the improved performance (+1.20\% AMOTA than the second BEVFusion method).}

Detector performance can also directly affect the tracking-by-detection framework. The universality of our CAMO-MOT method for different detectors is verified using a series of experiments involving multiple different detectors on both KITTI and nuScenes datasets, as shown in Table~\ref{table:3a} and Table~\ref{table:11a}. On the KITTI tracking validation set, we select four famous detectors including SECOND~\cite{lang2019pointpillars}, PVRCNN~\cite{shi2021pvrcnn}, PointRCNN~\cite{shi2019pointrcnn} and PointGNN~\cite{shi2020pointgnn} to evaluate our method. The results in Table~\ref{table:3a} illustrate that our method can achieve certain tracking performances on all detectors. When using the detected results from PointGNN~\cite{shi2020pointgnn}, our method achieves the best accuracy with a HOTA of 80.74\% for the Car class and 50.24\% for the Pedestrian class. On the nuScenes tracking validation set, we employ multiple 3D detectors including CenterPoint~\cite{yin2021center}, BEVFuison~\cite{https://doi.org/10.48550/arxiv.2205.13542}, FocalsConv~\cite{focalsconv-chen} to test, as shown in Table~\ref{table:11a}. We also show some other methods to compare. Using the same detector (CenterPoint) as other methods, our method maintains a strong performance in tracking. When using the fusion of BEVFuison~\cite{https://doi.org/10.48550/arxiv.2205.13542} and FocalsConv~\cite{focalsconv-chen} as the detector, our method can also achieve the best accuray with 76.3\% AMOTA. In addition, our method obtains minimal IDS without losing AMOTA precision, which illustrates that our method can prevent false matches to a large extent. These experiments effectively verify that our method can adapt to different detectors and demonstrate its universality.

\begin{table*}[t]
\caption[l]{
{A comparison of existing algorithms applied to the nuScenes tracking benchmark test set.}}
\label{table:10a}
\setlength\tabcolsep{9pt}
\centering
\begin{tabular}{lllllll}
\hline
\bf{Method}     & \bf{Detector}   & \bf{Modality}    & \bf{AMOTA$\uparrow$}   & \bf{FP$\downarrow$}    & \bf{FN$\downarrow$} & \bf{IDS$\downarrow$}  \\
\hline
EagerMOT~\cite{kim2021eagermot}  & CenterPoint~\cite{yin2021center}\&Cascade R-CNN~\cite{cai2018cascade} & LiDAR\&Camera   & 67.7 & 17705   & 24925  & 1156 \\
SimpleTrack~\cite{pang2021simpletrack}  & CenterPoint~\cite{yin2021center} & LiDAR  & 66.8  & 17514   & 23451  & 575 \\
CenterPoint~\cite{yin2021center}   & CenterPoint~\cite{yin2021center} & LiDAR & 65.0 & 17355   & 24557  & 684 \\
OGR3MOT~\cite{zaech2022learnable}   & CenterPoint~\cite{yin2021center} & LiDAR & 65.6 & 17877   & 24013  & \textbf{288} \\
CBMOT~\cite{benbarka2021score}   & CenterPoint~\cite{yin2021center}\&CenterTrack~\cite{2020Tracking} & LiDAR\&Camera & 68.1 & 21604   & 22828  & 709 \\
TransFusion~\cite{bai2021pointdsc}  &TransFuison~\cite{bai2021pointdsc} & LiDAR\&Camera & 71.8 & 16232   & 21846  & 944 \\
BEVFusion~\cite{https://doi.org/10.48550/arxiv.2205.13542}  & BEVFusion-e~\cite{https://doi.org/10.48550/arxiv.2205.13542} & LiDAR\&Camera & 74.1 & 19997   & 19395  & 506 \\
\hline
Ours      & BEVFuison~\cite{https://doi.org/10.48550/arxiv.2205.13542}\&FocalsConv~\cite{focalsconv-chen} & LiDAR\&Camera   & \bf{75.3} & \textbf{17269}  & \textbf{18192}  & 324  \\
\hline
\end{tabular}
\end{table*}

\begin{table*}[t]
\caption[l]{
{A comparison of existing algorithms applied to the nuScenes tracking benchmark val set. The numbers marked with~\cite{pang2021simpletrack} are from SimpleTrack~\cite{pang2021simpletrack}. }}
\label{table:11a}
\setlength\tabcolsep{9pt}
\centering
\begin{tabular}{lllllll}
\hline
\bf{Method}     & \bf{Detector}   & \bf{Modality}    & \bf{AMOTA$\uparrow$}   & \bf{AMOTP$\downarrow$}   & \bf{IDS$\downarrow$}  \\
\hline
AB3DMOT~\cite{weng2020ab3dmot, pang2021simpletrack}  & CenterPoint~\cite{yin2021center} & LiDAR  & 59.8 & 77.1     & 1570 \\
EagerMOT~\cite{kim2021eagermot}  & CenterPoint~\cite{yin2021center}\&Cascade R-CNN~\cite{cai2018cascade} & LiDAR\&Camera   & 71.2 & 56.9     & 899 \\
SimpleTrack~\cite{pang2021simpletrack}  & CenterPoint~\cite{yin2021center} & LiDAR   & 69.6  & 54.7   & 405 \\
CenterPoint~\cite{yin2021center}   & CenterPoint~\cite{yin2021center} & LiDAR & 66.5 & 56.7   & 562 \\
OGR3MOT~\cite{zaech2022learnable}   & CenterPoint~\cite{yin2021center} & LiDAR & 69.3 & 62.7    & 262 \\
CBMOT~\cite{benbarka2021score}   & CenterPoint~\cite{yin2021center}\&CenterTrack~\cite{2020Tracking} & LiDAR\&Camera & 69.2 & 56.3    & 459 \\
\hline
Ours      & CenterPoint~\cite{yin2021center} & LiDAR\&Camera   & 71.9 & \textbf{52.1}  & 393  \\
Ours      & BEVFuison~\cite{https://doi.org/10.48550/arxiv.2205.13542} & LiDAR\&Camera   & 76.0 & 56.1   & 243  \\
Ours      & FocalsConv~\cite{focalsconv-chen} & LiDAR\&Camera   & 75.3 & 52.7  & 337  \\
Ours      & BEVFuison~\cite{https://doi.org/10.48550/arxiv.2205.13542}\&FocalsConv~\cite{focalsconv-chen} & LiDAR\&Camera   & \textbf{76.3} & 52.7  & \textbf{239}  \\

\hline
\end{tabular}
\end{table*}

\subsection{Ablation Studies}

Ablation studies are also implemented to verify the effects of each modality on our proposed CAMO-MOT algorithm. Experiments involve images, point clouds, fusion methods, and fusion with an occlusion head applied to the KITTI validation set, and the results are shown in Table~\ref{table:4a}. In the case of the Car class, the fusion method (AP+MO) achieves a HOTA of +2.26\%, a MOTA of +0.33\%, and an IDS of -46 compared with the algorithm relying solely on point clouds (MO). It also achieves a HOTA of +19.13\%, a MOTA of +11.07\%, and an IDS of -17 compared with the algorithm relying solely on images (AP). Adding an occlusion head (AP+MO+OCC) produces a HOTA of +1.46\%, a MOTA of +0.71\%, and an IDS of -16 compared with the original fusion method (AP+MO). In the Pedestrian class, the fusion method (AP+MO) achieves a HOTA of +1.49\%, a MOTA of +0.14\%, and an IDS of -4 compared with point clouds (MO) and a HOTA of +13.69\%, a MOTA of +14.38\%, and an IDS of -52 compared with images (AP). The addition of an occlusion head (AP+MO+OCC) produces a HOTA of +0.48\%, a MOTA of +0.32\%, and an IDS of -8, compared with the original fusion method (AP+MO). These experiments show that our proposed fusion method achieves the best results and validates the role of the occlusion branch.

\begin{table}[t]
\caption{Ablation studies with varying occlusion states and appearance features in the KITTI tracking benchmark. AP refers to the appearance module; MO refers to the motion module; and OCC refers to the occlusion module.}
\label{table:4a}
\setlength\tabcolsep{5pt}
\begin{tabular}{llllll}
\hline
\bf{Type} &\bf{Module}       & \bf{Modality}                            & \bf{HOTA$\uparrow$} & \bf{MOTA$\uparrow$} & \bf{IDS$\downarrow$} \\ \hline
\multirow{4}{*}{Car}&AP              &Camera                             &  60.15    &  76.00      &  39  \\
&MO              & LiDAR        &  77.02    &  86.74   &  68   \\
&AP+MO           & LiDAR\&Camera                            & 79.28    &  87.07   &  22   \\
&AP+MO+OCC &  LiDAR\&Camera                           &  \bf{80.74}    &  \bf{87.78}   &  \bf{6} \\ \hline
\multirow{4}{*}{Ped.}&AP              & Camera                            &  36.07    &  42.49   &  148  \\
&MO              &LiDAR         &  48.27    &  56.73   &  100\\
&AP+MO           &  LiDAR\&Camera                           &  49.76    &  56.87   &  96  \\
&AP+MO+OCC &    LiDAR\&Camera                         &  \bf{50.24}    &  \bf{57.19}   &  \bf{88}\\ \hline
\end{tabular}
\end{table}

\begin{table}[h]
\centering
\caption{The results of ablation studies involving appearance features in the KITTI validation set. Compared variables include the optimal occlusion (OCC), features in the last three frames (LTF), and the method for weighted judging by occlusion (LTF+OCC).}
\label{table:7a}
\setlength\tabcolsep{5pt}
\begin{tabular}{llllllll}
\hline
\bf{Variable} & \bf{Type}                        & \bf{HOTA$\uparrow$}  & \bf{MOTA$\uparrow$}  & \bf{MOTP$\uparrow$}  & \bf{IDS$\downarrow$} & \bf{FP$\downarrow$}   & \bf{FN$\downarrow$}   \\\hline
OCC      & \multirow{3}{*}{Car}        & \bf{80.74} & \bf{87.78} & \bf{87.99} & \bf{6}   & \bf{544}  & \bf{474}  \\
LTF      &                             & 79.28 & 87.07 & 87.96 & 22   & 544  & 474  \\
LTF+OCC  &                             & 80.71 & 87.77 & 87.99 & 7   & 544  & 474  \\\hline
OCC      & \multirow{3}{*}{Ped.} & \bf{50.24} & \bf{57.19} & \bf{66.64} & \bf{88}  & \bf{2850} & \bf{1256} \\
LTF      &                             & 49.76 & 56.89 & 66.62 & 96  & 2850 & 1256 \\
LTF+OCC  &                             & 50.23 & 57.15 & 66.62 & 88  & 2850 & 1256 \\\hline
\end{tabular}
\end{table}

\begin{table}[t]
\centering
\caption{The results of ablation studies involving association rules. AP refers to the appearance module and MO refers to the motion module.}
\label{table:8a}
\setlength\tabcolsep{5pt}
\begin{tabular}{llllllll}
\hline
\bf{Variable} & \bf{Type}                        & \bf{HOTA$\uparrow$}  & \bf{MOTA$\uparrow$}  & \bf{MOTP$\uparrow$}  & \bf{IDS$\downarrow$} & \bf{FP$\downarrow$}   & \bf{FN$\downarrow$}   \\\hline
AP$\rightarrow$ MO    & \multirow{2}{*}{Car}        & 80.39 & 87.60 & 87.98 & 21  & 544  & 474  \\
MO$\rightarrow$ AP    &                              & \bf{80.74} & \bf{87.78} & \bf{87.99} & \bf{6}   & \bf{544}  & \bf{474}  \\\hline
AP$\rightarrow$ MO    & \multirow{2}{*}{Ped.} & 49.69 & 56.81 & 66.59 & 115 & 2853 & 1259 \\
MO$\rightarrow$ AP    &                             & \bf{50.24} & \bf{57.19} & \bf{66.64} & \bf{88}  & \bf{2850} & \bf{1256} \\\hline
\end{tabular}
\end{table}

\begin{table}[t]
\centering
\caption{The results of ablation studies involving category loss. NCL means no category loss and CL means category loss.}
\label{table:9a}
\setlength\tabcolsep{5pt}
\begin{tabular}{llllllll}
\hline
\bf{Variable}         & \bf{Type}                        & \bf{HOTA$\uparrow$}  & \bf{MOTA$\uparrow$}  & \bf{MOTP$\uparrow$}  & \bf{IDS$\downarrow$} & \bf{FP$\downarrow$}   & \bf{FN$\downarrow$}   \\\hline
NCL & \multirow{2}{*}{Car}        & 79.81 & 87.70 & 87.99 & 13  & 544  & 474  \\
CL    &                              & \bf{80.74} & \bf{87.78} & \bf{87.99} & \bf{6}   & \bf{544}  & \bf{474}  \\\hline
NCL & \multirow{2}{*}{Ped.} & 49.99 & 56.76 & 66.55 & 114 & 2853 & 1259 \\
CL    &                             & \bf{50.24} & \bf{57.19} & \bf{66.64} & \bf{88}  & \bf{2850} & \bf{1256} \\\hline
\end{tabular}
\end{table}

\indent To verify that our chosen optimal occlusion features are valid, we separately calculate (1) the features based on optimal occlusion situations (OCC), (2) the average value of appearance features for the last three frames (LTF), (3) the weighted accumulation of appearance features for the last three frames according to the occlusion situation, as shown in Table~\ref{table:7a}. The resulting occlusion weights are 1, 0.7, 0.3, and 0 according to occlusion state (fully visible, partly occluded, largely occluded, fully occluded), respectively. It is evident that tracking effects are best in the case of OCC. It can be seen from the results that whether OCC or LTF+OCC, it is better than LTF, indicating that the occlusion head plays an important role in our method. When introducing the object appearance of multiple frames, due to the different occlusion states, interference of different degrees will be introduced, so that the final appearance features of the object cannot fully represent the object itself and the performance of LTF+OCC is not as good as that of OCC.

We also test algorithm performance under the two association rules of appearance followed by movement and then movement followed by appearance, as shown in Table~\ref{table:8a}. This MO$\rightarrow$AP and AP$\rightarrow$MO approach produces HOTA values of +0.35\% and +0.55\%, MOTA values of +0.18\% and +0.38\% and IDS values of -15 and -27 for the Car and Pedestrian categories, respectively. We analyze that the possible reason is that when there are objects with very similar appearances in the scene, identity switch is likely to occur, resulting in a decrease in the metrics.

\indent The effectiveness of category loss during tracking is also evaluated experimentally. As shown in Table~\ref{table:9a}, the HOTA of Car and Pedestrian classes in the method with category loss (CL) increases +0.93\% and +0.25\% than the one with no category loss (NCL). This suggests our proposed method eliminates the possibility of associations between categories effectively, significantly increasing the tracking accuracy for Car and Pedestrian classes.

\subsection{Visualization}

\begin{figure*}[ht]
\begin{center}
\includegraphics[width=1\textwidth,height=0.8\textwidth]{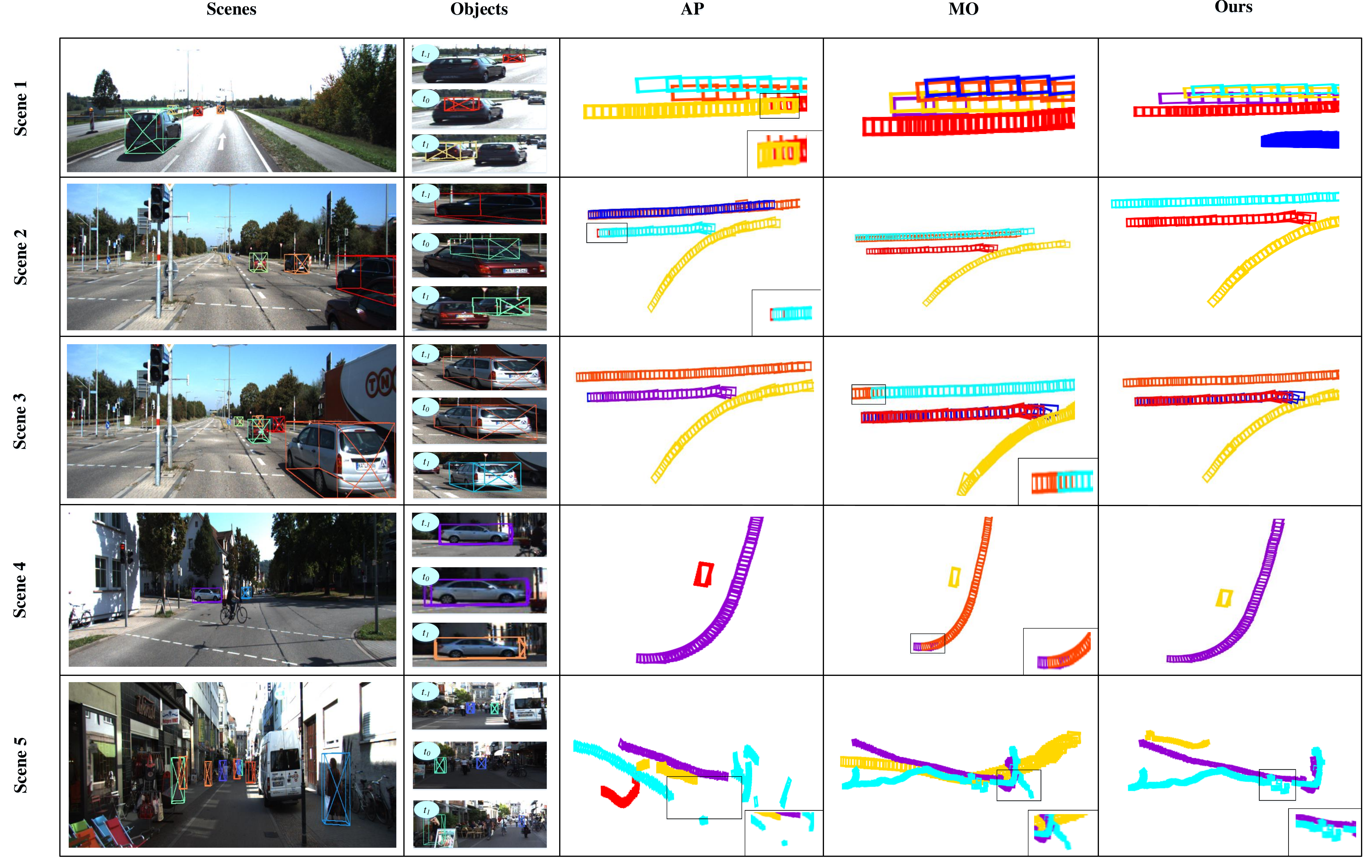}
\caption{Qualitative visualization of object trajectories with single and fusion modalities. AP and MO refer to the appearance module and motion module, respectively, while the ``Scenes'' and ``Objects'' columns provide corresponding image scenes and detected objects. The $t_{-1}$, $t_{0}$, and $t_{1}$ terms denote the previous, current, and next moments, respectively. The third, fourth and fifth columns present the 2D object trajectories and colors mean the different detected objects. Our method achieves the lowest ID switches.}
\label{fig:f4}
\end{center}
\end{figure*}

The effectiveness of the fusion method is further evaluated using several qualitative visualization experiments on the KITTI dataset. 
\\\indent The results of a tracking method using single and fusion modalities are provided in Fig.~\ref{fig:f4}. In scenes 1 and 2, the tracked object is considered a new entity due to occlusions in the movement process. However, our method allows for stable tracking without occlusion effects. In scene 3, the tracker considers the object to be a new object when it suddenly accelerates. In scene 4, the object is considered a new one because of a sudden turn. None of these effects are observed using our method. In scene 5, the object moves irregularly and is also occluded by other objects. Although our 
method can not resolve this situation entirely, it provides obvious improvement compared with single modalities.

\begin{figure*}[t]
\begin{center}
\includegraphics[width=0.9\textwidth]{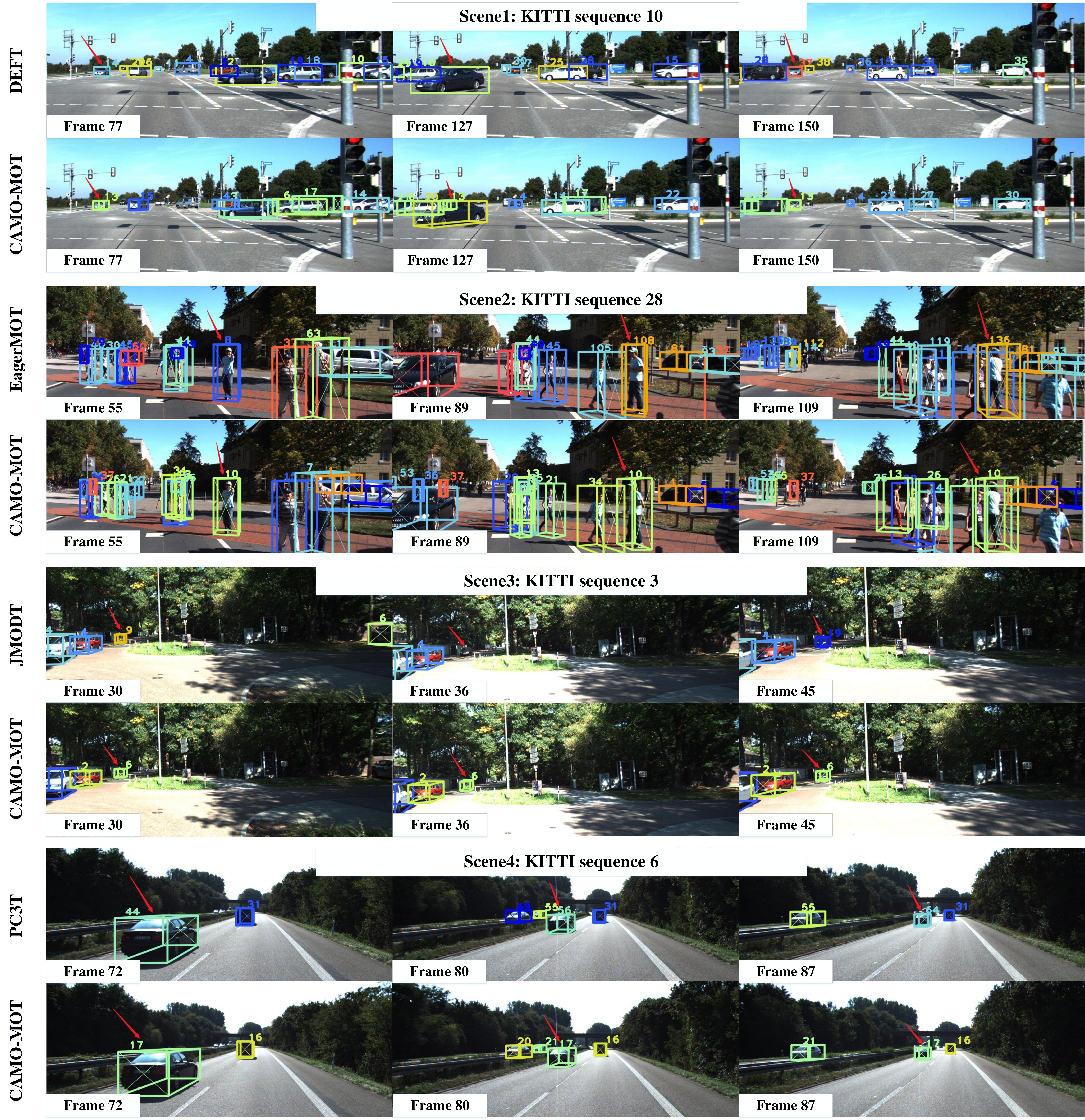}
\caption{Qualitative results comparison between CAMO-MOT and DEFT~\cite{chaabane2021deft}, EagerMOT~\cite{kim2021eagermot}, JMODT~\cite{huang2021joint}, PC3T~\cite{20213D} on KITTI tracking benchmark. Each pair of rows shows the comparison of the results for one sequence. The color of the boxes represents the identity of the tracks. Red arrows point at tracking failures (identity switches). Under occlusion situations, wrong associations occur: DEFT (ID 14$\rightarrow$ID 37), EagerMOT (ID 8$\rightarrow$ID 136). When the illumination changes, JMODT shows the wrong ID switch (ID 9$\rightarrow$ID 19). When the object is accelerating, ID 44$\rightarrow$ID 64 occurs in PC3T. However, our CAMO-MOT can still keep stable tracking in all these situations}
\label{fig:cp}
\end{center}
\end{figure*}

\indent We also provide qualitative results comparison between our CAMO-MOT and DEFT~\cite{chaabane2021deft}, EagerMOT~\cite{kim2021eagermot}, JMODT~\cite{huang2021joint}, PC3T~\cite{20213D} on KITTI tracking benchmark, as shown in Fig.~\ref{fig:cp}. Compared with DEFT and EagerMOT in occlusion, they both have identity switches, but our method can achieve stable tracking. PC3T utilizes only point cloud data and also has identity switches when the object continuously accelerates, but our method is still stable. When the illumination changes, the multi-modal fusion method JMODT recognizes the tracking object as a new one while our method still maintains stable tracking.

\indent The above comparisons illustrate that our method can utilize the information of the camera and LiDAR to achieve stable tracking effectively, even in severe occlusions.


\section{Conclusion}
A new multi-modal 3D MOT framework CAMO-MOT based on the combined appearance-motion optimization is proposed to fuse camera and LiDAR information and achieve stable tracking effectively. An occlusion head is designed to identify object occlusion states and select optimal appearance features to reduce the effects of occlusions. We also propose a 3D motion module based on confidence scores to eliminate false detections. This study represents the first attempt at introducing category loss to ensure objects are only related within the same category. Our method achieves the state-of-the-art performance among multi-modal MOT algorithms applied to the KITTI tracking benchmark and the state-of-the-art performance among all MOT algorithms applied to the nuScenes tracking benchmark.
\textcolor{black}{In addition, it achieves the lowest IDS among all algorithms on the KITTI test set and among the top ten algorithms on the nuScenes test set, indicating remarkable safety and stability.} 

\section{Acknowledgments}
We thank LetPub (www.letpub.com) for linguistic assistance and pre-submission expert review.

{\small
\bibliographystyle{IEEEtran}
\bibliography{IEEEabrv,CAMOMOT_with_nuScenes_arXiv}
}

\end{document}